\documentclass[pdflatex,sn-basic]{sn-jnl}

\usepackage{graphicx}%
\usepackage{multirow}%
\usepackage{amsmath,amssymb,amsfonts}%
\usepackage{amsthm}%
\usepackage{mathrsfs}%
\usepackage[title]{appendix}%
\usepackage{xcolor}%
\usepackage{textcomp}%
\usepackage{manyfoot}%
\usepackage{booktabs}%
\usepackage{algorithm}%
\usepackage{algorithmicx}%
\usepackage{algpseudocode}%
\usepackage{listings}%
\usepackage{subcaption}

\begin{document}

\title[HAMDPO]{Heterogeneous Multi-Agent Reinforcement Learning via Mirror Descent Policy Optimization}

\author[1]{\fnm{Mohammad Mehdi} \sur{Nasiri }}\email{mehdinasiri@modares.ac.ir}

\author*[1]{\pfx{} \fnm{Mansoor} \sur{Rezghi}}\email{rezghi@modares.ac.ir}

\affil[1]{\orgdiv{Department of Computer Science}, \orgname{Tarbiat Modares University}, \orgaddress{
\city{Tehran},
\state{Tehran},
\country{Iran}}}


\abstract{This paper presents an extension of the Mirror Descent method to overcome challenges in cooperative Multi-Agent Reinforcement Learning (MARL) settings, where agents have varying abilities and individual policies. The proposed Heterogeneous-Agent Mirror Descent Policy Optimization (HAMDPO) algorithm utilizes the multi-agent advantage decomposition lemma to enable efficient policy updates for each agent while ensuring overall performance improvements. By iteratively updating agent policies through an approximate solution of the trust-region problem, HAMDPO guarantees stability and improves performance. Moreover, the HAMDPO algorithm is capable of handling both continuous and discrete action spaces for heterogeneous agents in various MARL problems. We evaluate HAMDPO on Multi-Agent MuJoCo and StarCraftII tasks, demonstrating its superiority over state-of-the-art algorithms such as HATRPO and HAPPO. These results suggest that HAMDPO is a promising approach for solving cooperative MARL problems and could potentially be extended to address other challenging problems in the field of MARL.}


\maketitle

\section{Introduction} 
Multi-agent Reinforcement Learning (MARL) is an important tool to solve a wide range of sequential decision-making problems in real-world applications, such as autonomous driving \citep{zhou2020smarts}, traffic signal control \citep{wang2020large}, drone swarms \citep{boubin2021programming}, and smart grid \citep{roesch2020smart}. In these problems, multiple agents must work together to efficiently accomplish a task by achieving maximum reward in a shared environment. However, developing algorithms to outperform single-agent algorithms in such settings requires dealing with additional challenges such as scalability and non-stationarity.

Cooperative MARL involves several agents learning to jointly accomplish a task. Although several paradigms have been introduced to employ single-agent training methods, such as policy gradient in MARL \citep{lowe2017multi,foerster2018counterfactual}, scaling up single-agent methods to multi-agent systems presents enduring challenges \citep{gupta2017cooperative}. Thus, researchers continue to develop new algorithms that can effectively address these challenges and improve the performance of cooperative MARL. One of the earliest methods for MARL is called Decentralized Training with Decentralized Execution (DTDE) \citep{Tan1993MultiAgentRL}. In this paradigm, each agent makes decisions and learns independently, without access to other agents' observations, actions, or policies. This method is easy to implement, but as the number of agents increases, learning becomes unstable because agents do not take into account other agents' behavior during policy updates. Therefore, the environment appears non-stationary from a single agent's viewpoint.

To improve the stability of the learning process, another method called Centralized Training with Centralized Execution (CTCE) was introduced \citep{gupta2017cooperative}. This method models the joint policy as a centralized policy and uses single-agent methods to learn this aggregated policy. However, scalability suffers from the curse of dimensionality because the state-action space grows exponentially with the number of agents. To address this issue, the joint policy can be factored into individual policies for each agent. Centralized Training with Decentralized Execution (CTDE) \citep{lowe2017multi} leverages the benefits of both DTDE and CTCE. In CTDE, each agent holds an individual policy and learns through actor-critic \citep{mnih2016asynchronous} or policy gradient methods. The critic model has access to the global state and other agents' actions, so it learns the true state-value. When paired with the actor, it can be used to find the optimal policy. CTDE allows agents to use this additional information during training, but in execution time, each agent acts independently and decision-making relies only on its own policy. Therefore, CTDE is a promising solution for many real-world AI applications. 

On the other hand, in Reinforcement Learning (RL), sudden changes in policy distribution can lead to poor local optima. To avoid this, in single-agent, trust-region algorithms guarantee that policy updates are not excessively drastic and the new policy remains close to the old one, ensuring monotonic improvement in policy updates. \citealp{kakade2002approximately} derived a policy improvement bound and introduced the Conservative Policy Iteration (CPI) algorithm to stabilize and enhance the policy learning process. \citealp{schulman2015trust} extended this policy improvement bound for stochastic policies and introduced Trust Region methods such as trust-region Policy Optimization (TRPO) \citep{schulman2015trust} and Proximal Policy Optimization (PPO) \citep{schulman2017proximal} algorithms to solve this trust-region problem. These algorithms have been shown to be effective in a wide range of tasks, including robotic control and game-playing.

The CTDE provides a framework for extending policy gradient theorems \citep{sutton2018reinforcement} from single-agent to multi-agent scenarios. One example of a policy gradient method that utilizes separate actors and critics for each agent and trains the critic in a centralized way is Multi-Agent Deep Deterministic Policy Gradient \citep{lowe2017multi}. While several policy gradient approaches are available, trust-region methods like the TRPO and PPO algorithms are state-of-the-art single-agent policy gradient methods. Attempts to extend trust-region learning to multi-agent settings include Independent PPO (IPPO) \citep{de2020independent} and Multi-Agent PPO (MAPPO) \citep{yu2021surprising}. However, these methods suffer from the drawbacks of homogeneous agents and parameter sharing. Homogeneous agents share a common set of skills, enabling them to share parts of their learning model with other agents and use the same policy network. Parameter sharing can accelerate the learning progress of this network. However, in scenarios where agents are heterogeneous and have different tasks, their action space may not be identical, which can limit the effectiveness of parameter sharing.

Recently, \citealp{kuba2022trust} proposed the Heterogeneous-Agent TRPO (HATRPO) and Heterogeneous-Agent PPO (HAPPO) algorithms to address limitations of existing trust-region learning methods in MARL, such as IPPO and MAPPO. HATRPO/HAPPO rely on the multi-agent advantage decomposition lemma and sequential policy update scheme to ensure monotonic improvement. The outstanding HATRPO/HAPPO algorithms are extensions of single-agent TRPO/PPO methods designed for MARL with heterogeneous agents.

Although HATRPO/HAPPO algorithms showed appropriate results, recently, a single-agent policy gradient method called Mirror Descent Policy Optimization (MDPO) was introduced by \citealp{tomar2020mirror}. MDPO has shown superior performance compared to TRPO and PPO in single-agent scenarios. Therefore, extending the MDPO algorithm to MARL with heterogeneous agents would enable us to leverage the advantages of mirror descent in multi-agent settings, which could work better than HATRPO/HAPPO.

This paper introduces a novel trust-region algorithm for MARL called Heterogeneous-Agent Mirror Descent Policy Optimization (HAMDPO). HAMDPO leverages the multi-agent advantage decomposition lemma and the sequential policy update scheme to apply mirror descent in multi-agent scenarios. We evaluate HAMDPO on several tasks of the StarCraftII and Multi-Agent Mujoco continuous benchmark and compare its performance with that of HATRPO and HAPPO. Our results demonstrate that HAMDPO outperforms both HATRPO and HAPPO in terms of convergence and overall performance. 

This paper is organized as follows: Section \ref{sec_Preliminaries} provides Preliminaries, offering an overview of essential background concepts. Section \ref{sec_MD&PO} explores Mirror Descent and Policy Optimization in multi-agent systems. Section \ref{sec_MATRL} presents Multi-Agent Trust-Region Learning, addressing trust and cooperation challenges, enhancing agent interactions' stability and performance. Section \ref{sec_HAMDPO} introduces Heterogeneous-Agent Mirror Descent Policy Optimization, handling agent heterogeneity to improve system performance. Section \ref{sec_Experiments} discusses Experiments, comprehensively evaluating and comparing the effectiveness of our approaches.

\section{Preliminaries}\label{sec_Preliminaries}
In a fully cooperative multi-agent setting, we consider a Markov game \citep{littman1994markov} represented by a tuple $\langle \mathcal{N}, \mathcal{S}, \boldsymbol{\mathcal{A}}, P, r, \gamma \rangle$, where $\mathcal{N} = {1,\ldots,n}$ denotes the set of $n$ interacting agents, $\mathcal{S}$ is the set of states observed by all agents, and $\boldsymbol{\mathcal{A}} = \mathcal{A}^1 \times \cdots \times \mathcal{A}^n$ denotes the joint action space of $n$ agents. Here, $\mathcal{A}^i$ is the action space of agent $i$. The transition probability from any state $s \in \mathcal{S}$ to any state $s' \in \mathcal{S}$ for any joint action $\boldsymbol{a} \in \boldsymbol{\mathcal{A}}$ is given by $P: \mathcal{S} \times \boldsymbol{\mathcal{A}} \times \mathcal{S} \rightarrow [0,1]$. The reward function is $r: \mathcal{S} \times \boldsymbol{\mathcal{A}} \rightarrow \mathbb{R}$, and $\gamma \in [0,1)$ is the discount factor.

At time step $t$, each agent $i$ chooses an action $a^i_t$ according to its policy $\pi^i$ given the current state $s_t$. The agents joint action is then formed as $\boldsymbol{a} = (a^1, \ldots, a^n)$, drawn from the joint policy $\boldsymbol{\pi} = (\pi^1 \times \cdots \times \pi^n)$. The agents then receive a reward $r_t$ and move to the next state $s_{t+1}$.

The joint value function in a state $s$ is defined as $V_{\boldsymbol{\pi}}(s) \triangleq \mathbb{E}_{\mathbf{a}_{0: \infty} \sim \boldsymbol{\pi}, \mathrm{s}_{1: \infty} \sim P}\left[\sum_{t=0}^{\infty} \gamma^{t} \mathrm{r}_{t} \mid \mathrm{s}_{0}=s\right]$, which is the expected cumulative reward obtained by following the joint policy $\boldsymbol{\pi}$ from state $s$. The joint state-action value function $Q_{\boldsymbol{\pi}}(s, \boldsymbol{a}) \triangleq \mathbb{E}_{\mathrm{s}_{1: \infty} \sim P, \mathbf{a}_{1: \infty} \sim \boldsymbol{\pi}}\left[\sum_{t=0}^{\infty} \gamma^{t} \mathrm{r}_{t} \mid \mathrm{s}_{0}=s, \mathbf{a}_{0}=\boldsymbol{a}\right]$ represents the expected cumulative reward obtained by following the joint policy $\boldsymbol{\pi}$ from state $s$ and taking joint action $\boldsymbol{a}$. The difference between these two functions gives the joint advantage function $A_{\boldsymbol{\pi}}(s, \boldsymbol{a}) \triangleq Q_{\boldsymbol{\pi}}(s, \boldsymbol{a})-V_{\boldsymbol{\pi}}(s)$.

In a fully-cooperative multi-agent setting, agents collaborate to maximize cumulative reward. This is achieved by maximizing the expected sum of discounted rewards obtained by the agents over
time denoted by the following objective function:
\begin{equation}
\mathcal{J}(\boldsymbol{\pi}) \triangleq \mathbb{E}_{\mathrm{s}_{0: \infty} \sim \rho_{\boldsymbol{\pi}}^{0: \infty}, \mathbf{a}_{0: \infty} \sim \boldsymbol{\pi}}\left[\sum_{t=0}^{\infty} \gamma^{t} \mathrm{r}_{t}\right]
\end{equation}
Here, the expectation is taken over the state-action marginal distribution of the trajectory induced by the joint policy $\boldsymbol{\pi}$, denoted by $\rho_{\boldsymbol{\pi}}$. 

It should be noted that in the fully-cooperative setting, agents work together towards a common goal and do not have conflicting objectives. This is in contrast to the partially cooperative or non-cooperative settings, where agents may have individual objectives that may not align with the collective goal \citep{oroojlooy2023review}.

\subsection{Mirror Descent and Policy Optimization} \label{sec_MD&PO}
Trust-region methods have demonstrated their effectiveness in finding optimal policies and stabilizing learning in RL. Notably, the TRPO and PPO algorithms are recognized as remarkable trust-region approaches. 
Despite the extensive study of the mirror descent algorithm in tabular single-agent RL \citep{shani2020adaptive, geist2019theory}, where \citealp{shani2020adaptive} proved a convergence rate of $\tilde{O}(1/\sqrt{K})$ for MD-style RL algorithms in the tabular case, a significant advancement was made by \citealp{tomar2020mirror} with the introduction of MDPO. This extension enables the training of parametric policies using neural networks and has consistently outperformed TRPO and PPO in a wide range of tasks.

To solve a trust-region problem in MARL using the mirror descent algorithm, we first need to introduce the mirror descent definition and its utilization in the RL setting. Consider a constrained convex problem of the form:
\begin{equation} 
x^{*} \in \underset{x \in C}{\arg \min } , f(x), \label{eq:mdpofx}
\end{equation}
where $f$ is a convex function and $C$ is a convex and compact constraint set. The mirror descent algorithm \citep{beck2003mirror} with step size $t_k$ solves problem (\ref{eq:mdpofx}) by using the first-order approximation of $f(x)$, augmented with a regularization term, as follows:
\begin{equation}
x_{k+1} \in \underset{x \in C}{\arg \min }\left\langle\nabla f\left(x_{k}\right), x-x_{k}\right\rangle+\frac{1}{t_{k}} B_{\psi}\left(x, x_{k}\right),
\end{equation}
where $B_{\psi}\left(x, x_{k}\right)$ is the Bregman divergence associated with the Bregman potential $\psi$, defined as:
\begin{equation}
B_{\psi}\left(x, x_{k}\right)=\psi(x)-\psi\left(x_{k}\right)-\left\langle\nabla \psi\left(x_{k}\right), x-x_{k}\right\rangle.
\end{equation}

A strictly convex and twice differentiable function $\psi: C \rightarrow \mathbb{R}$ is called a Bregman potential on the convex domain $C$. The mirror descent algorithm uses a Bregman potential $\psi$ to obtain a local update rule that takes into account the global geometry of the constraint set.  By selecting the negative Shannon entropy as the Bregman potential, we obtain the KL-divergence between the probability distributions as the Bregman divergence.

In summary, mirror descent is a powerful optimization technique that can be used to solve constrained convex optimization problems in the context of reinforcement learning. By incorporating the Bregman divergence with a carefully chosen Bregman potential, mirror descent can account for the geometry of the constraint set and efficiently compute a solution to the optimization problem.

In single-agent policy gradient algorithms, the optimization problem of finding the optimal policy $\pi^{*}$ can be expressed as:
\begin{equation} 
\pi^{*} \in \underset{\pi \in \Pi}{\arg \max } \, \mathbb{E}_{s \sim \mu}\left[V^{\pi}(s)\right] , \label{eq:rlobj}
\end{equation}
where $\mu$ represents the initial state distribution and  $\Pi=\{\pi(\cdot \mid s ; \theta): s \in \mathcal{S}, \theta \in \Theta\}$ is a class of smoothly parameterized stochastic policies with parameter $\theta$, which is also used to represent a policy.

Previous works have studied mirror descent in both the tabular and parametric cases. They have shown that mirror descent can be used to solve the optimization problem of policy gradient methods and can yield excellent results even with non-convex objective functions. The on-policy update rule for mirror descent is defined as:
\begin{equation} 
\pi_{k+1} \leftarrow \underset{\pi \in \Pi}{\arg \max } \, \mathbb{E}_{s \sim \rho_{\pi_{k}}}\left[\mathbb{E}_{a \sim \pi}\left[A^{\pi_{k}}(s, a)\right]-\frac{1}{t_{k}} \mathrm{KL}\left( \pi, \pi_{k}\right)\right], \label{eq:1}
\end{equation}
where $\rho_{\pi_{k}}$ is the state distribution of the previous policy \citep{tomar2020mirror}. However, taking only one step of stochastic gradient descent (SGD) over the update rule (\ref{eq:1}) leads to the gradient of the KL term being equal to zero i.e. $\left.\nabla_{\theta} \mathrm{KL}\left(\pi_{\theta}, \pi_{\theta_{k}}\right)\right|_{\theta=\theta_{k}}=0$, which reduces to the vanilla policy gradient method \citep{sutton1999policy}. Therefore, multiple SGD steps are needed at each iteration to account for the regularization term. Because samples are drawn from the previous policy distribution and the policy distribution changes with every SGD step, so importance sampling is employed to correct the estimation. Finally, the MDPO objective is defined as:
\begin{align}
L\left(\theta,\theta_k\right) &=  \mathbb{E}_{s \sim \rho_{\pi_{\theta_{k}}}}\left[\mathbb{E}_{a \sim \pi_{\theta_{k}}}\left[\frac{\pi_{\theta}(a \mid \mathrm{s})}{ \pi_{\theta_{k}}(a \mid \mathrm{s})}A^{\pi_{\theta_{k}}}(s, a)\right]- \frac{1}{t_{k}} \mathrm{KL}\left( \pi_{\theta}(\cdot \mid \mathrm{s}), \pi_{\theta_{k}}(\cdot \mid \mathrm{s})\right)\right]
\end{align}

To solve the policy optimization problem (\ref{eq:rlobj}), TRPO approximates the objective function using the first-order term and approximates the constraint using the second-order term of the corresponding Taylor series. This results in a policy gradient update that involves calculating the inverse of the Fisher information matrix, which is a quadratic approximation to the constraint. Despite the benefits of TRPO, it is a complicated and expensive algorithm to run.

To address these issues, PPO relaxes the hard constraint assumption and reduces the computation burden of TRPO. PPO attempts to simplify the optimization process while still retaining the advantages of TRPO. It uses a clipped surrogate objective and updates the policy by solving an unconstrained optimization problem in which the ratio of the new to old policy is clipped to remain bounded. Unfortunately, PPO violates the trust-region assumptions by failing to constrain the update size and does not guarantee monotonic improvement to be satisfied \citep{wang2020truly,Engstrom2020}.

The MDPO algorithm, unlike TRPO, is a first-order method and is more straightforward to implement. Additionally, MDPO effectively applies a trust-region constraint, which is an advancement over PPO. Since MDPO has demonstrated strong performance in the context of single-agent RL, we intend to investigate its applicability and effectiveness in MARL.

\section{Multi-Agent Trust-Region Learning} \label{sec_MATRL}

MARL is a sub-field of RL that deals with the problem of multiple agents interacting in the same environment. One of the challenges in MARL is designing algorithms that enable agents to learn from each other and cooperate effectively. One popular technique for achieving this is parameter sharing, which allows agents to share the same set of parameters for their policies or value functions.

Parameter sharing has several benefits. For one, it can reduce the number of parameters in a model, making it more efficient and easier to train. Additionally, by allowing agents to learn from each other's experiences, it can boost overall learning performance \citep{gupta2017cooperative}. However, there are also drawbacks to parameter sharing. For example, it can prevent agents from developing unique behaviors, which can limit their effectiveness in certain situations. Recent studies have also shown that parameter sharing can lead to a sub-optimal outcome that becomes exponentially worse as the number of agents increases \citep{kuba2022trust}.

Most of trust-region methods in MARL use parameter sharing and are designed for homogeneous agents and do not guarantee monotonic improvement \citep{de2020independent,yu2021surprising} . Recently, \citealp{kuba2022trust} developed a theoretically-justified trust-region learning framework for heterogeneous agents by introducing the multi-agent advantage decomposition lemma.  To use this framework, it is necessary to define state-action value and advantage functions for different subsets of agents. The multi-agent state-action value function for a subset $i_{1:m}$ of n agents and its complement $-i_{1:m}$ is defined as:
\begin{equation}
Q_{\boldsymbol{\pi}}^{i_{1: m}}\left(s, \boldsymbol{a}^{i_{1: m}}\right) \triangleq \mathbb{E}_{\mathbf{a}^{-i_{1: m}} \sim \boldsymbol{\pi}^{-i_{1: m}}}\left[Q_{\boldsymbol{\pi}}\left(s, \boldsymbol{a}^{i_{1: m}}, \mathbf{a}^{-i_{1: m}}\right)\right].
\end{equation}

Here, $\boldsymbol{\pi}$ is the joint policy of all agents, $\boldsymbol{a}^{i_{1:m}}$ is the joint action of agents in subset $i_{1:m}$, and $\mathbf{a}^{-i_{1:m}}$ is the joint action of agents in the complement subset $-i_{1:m}$. The expectation is taken over the joint action of agents in the complement subset, sampled from the joint policy of agents in the complement subset, $\boldsymbol{\pi}^{-i_{1:m}}$.

Correspondingly, the multi-agent advantage function for two disjoint subsets $i_{1: m}$ and $j_{1: k}$ is 
\begin{equation}
A_{\boldsymbol{\pi}}^{i_{1: m}}\left(s, \boldsymbol{a}^{j_{1: k}}, \boldsymbol{a}^{i_{1: m}}\right) \triangleq Q_{\boldsymbol{\pi}}^{j_{1: k}, i_{1: m}}\left(s, \boldsymbol{a}^{j_{1: k}}, \boldsymbol{a}^{i_{1: m}}\right)-Q_{\boldsymbol{\pi}}^{j_{1: k}}\left(s, \boldsymbol{a}^{j_{1: k}}\right) .
\end{equation}
For an arbitrary subset $i_{1: m}$ with a given joint policy $\boldsymbol{\pi}$ and state $s$, the multi-agent advantage decomposition lemma is 
\begin{equation}
\label{equ_adv_dec}
A_{\boldsymbol{\pi}}^{i_{1: m}}\left(s, \boldsymbol{a}^{i_{1: m}}\right)=\sum_{j=1}^{m} A_{\boldsymbol{\pi}}^{i_{j}}\left(s, \boldsymbol{a}^{i_{1: j-1}}, a^{i_{j}}\right).
\end{equation}

The equation (\ref{equ_adv_dec}) shows that the joint advantage function can be derived from a summation of each agent's local advantage. This means that for an arbitrary subset $i_{1: m}$ with a given joint policy $\boldsymbol{\pi}$ and state $s$, the advantage function can be expressed as the sum of each agent's advantage function for their actions. As a result, any agent that improves its own advantage will also improve the joint advantage. This is in contrast to VDN or QMIX \citep{sunehag2017value, rashid2018qmix}, which rely on assumptions about the decomposability of the joint value function. Therefore, the multi-agent advantage decomposition lemma is a powerful tool that allows agents to learn independently while still improving the overall performance.

Consider $\boldsymbol{\pi}$ as a joint policy, $\boldsymbol{\bar{\pi}^{i_{1:m-1}}}$ as some other joint policy of agents $i_{1:m-1}$, and $\hat{\pi}^{i_m}$ as some other policy of agent $i_m$. The surrogate objective is defined as follows:
\begin{equation}
	L_{\boldsymbol{\pi}}^{i_{1:m}}(\boldsymbol{\bar{\pi}}^{i_{1:m-1}},\hat{\pi}^{i_m}) = \mathbb{E}_{s\sim\rho_{\boldsymbol{\pi}},\boldsymbol{a}^{i_{1:m-1}}\sim \boldsymbol{\bar{\pi}^{i_{1:m-1}}},a^{i_{m}}\sim \hat{\pi}^{i_{m}}}\left[ A^{i_m}_{\boldsymbol{\pi}} \left( s,\boldsymbol{a}^{i_{1:m-1}},a^{i_{m}}\right) \right]
\end{equation}
where $A^{i_m}_{\boldsymbol{\pi}} \left( s,\boldsymbol{a}^{i_{1:m-1}},a^{i_{m}}\right)$ is the advantage function for agent $i_m$ with respect to the joint policy $\boldsymbol{\pi}$. Let $D_{\mathrm{KL}}^{\max }\left(\pi, \bar{\pi}\right) = \underset{s}{\max} \, D_{\mathrm{KL}}\left(\pi(.|s), \bar{\pi}(.|s)\right)$, the policy improvement bound for two joint policies $\boldsymbol{\pi}$ and $\boldsymbol{\bar{\pi}}$ is:
\begin{equation}
\mathcal{J}(\boldsymbol{\bar{\pi}}) \ge \mathcal{J}(\boldsymbol{\pi}) + \sum_{m=1}^{n}\left[ L_{\boldsymbol{\pi}}^{i_{1:m}}\left( \boldsymbol{\bar{\pi}}^{i_{1:m-1}},\bar{\pi}^{i_{m}} \right) - C D_{KL}^{max} \left( \pi^{i_m},\bar{\pi}^{i_m} \right) \right], \label{ineq:improvebound}
\end{equation}
where $\mathcal{J}(\boldsymbol{\pi})$ and $\mathcal{J}(\boldsymbol{\bar{\pi}})$ are the expected returns for joint policies $\boldsymbol{\pi}$ and $\boldsymbol{\bar{\pi}}$, respectively and the constant $C=\frac{4 \gamma \max {s, \boldsymbol{a}}\left|A{\boldsymbol{\pi}}(s, \boldsymbol{a})\right|}{(1-\gamma)^2}$.
Thus, any agent that updates its policy in a way that positively impacts the above summation is guaranteed to increase the joint policy performance  \citep{kuba2022trust}.

HATRPO/HAPPO are derived from the aforementioned bounds and update each agent's policy sequentially, rather than updating the entire joint policy at once. The sequential update follows the policy improvement bound (\ref{ineq:improvebound}), and each agent has a distinct optimization objective that takes into account the previous agents' updates. The TRPO/PPO update rule is utilized to optimize each agent's policy during the sequential update.

\section{Heterogeneous Agent Mirror Descent Policy Optimization} \label{sec_HAMDPO}
In the fully cooperative MARL setting, we consider agents that behave independently and have distinct policies, making them heterogeneous agents. This characteristic allows us to directly apply the MDPO algorithm to MARL using the trust-region learning framework proposed by \citealp{kuba2022trust}. Thus, we present a mirror descent objective to update the policies of these heterogeneous agents. The optimization problem for finding the optimal joint policy is defined as follows:
\begin{equation}
\boldsymbol{\pi}^{*} \in \underset{\boldsymbol{\pi} \in \boldsymbol{\Pi}}{\arg \max } , \mathbb{E}_{s \sim \mu}\left[V_{\boldsymbol{\pi}}(s)\right] ,
\end{equation}
where $\boldsymbol{\pi}$ represents the joint policy and $\boldsymbol{\Pi}$ denotes the set of all possible joint policies.
To solve this optimization problem and extend the mirror descent update rule to MARL, we can employ the joint advantage function within the HAMDPO agent objective function, which is based on trust-region theory in MARL \citep{kuba2022trust}. Let $\boldsymbol{\pi} = \prod_{i=1}^n \pi^i$ and $\overline{\boldsymbol{\pi}} = \prod_{i=1}^n \bar{\pi}^i$ be joint policies. The on-policy version of MDPO for updating the joint policy is derived as follows:
\begin{equation} 
\boldsymbol{\pi}_{k+1} \leftarrow \underset{\overline{\boldsymbol{\pi}} \in \boldsymbol{\Pi}}{\arg \max }  \;\mathbb{E}_{\mathrm{s} \sim \rho_{\boldsymbol{\pi}_{k}}, \mathbf{a} \sim \overline{\boldsymbol{\pi}}}\left[A_{\boldsymbol{\pi}_{k}}(\mathrm{s}, \mathbf{a})\right]-\frac{1}{t_{k}} \mathrm{KL}(\overline{\boldsymbol{\pi}}, \boldsymbol{\pi}_{k} ) \label{eq:jointMDPO}
\end{equation}

The update rule (\ref{eq:jointMDPO}) consists of two terms: the expected joint advantage function, which quantifies the performance improvement of the joint policy compared to the old policy, and the KL-divergence, which promotes the joint policy to stay close to the old policy to ensure stable learning. To accommodate heterogeneous agents using the CTDE approach, we can decompose both terms and derive an update rule by aggregating the local advantages and KL terms of the agents. This allows us to update the joint policy by taking into account the individual contributions of each agent.

To decompose the joint advantage function, we utilize the joint advantage decomposition lemma in equation (\ref{equ_adv_dec}). Additionally, for the KL-divergence term of two joint policies, following a similar approach to \citealp{kuba2022trust}, let $\boldsymbol{\pi}$ and $\overline{\boldsymbol{\pi}}$ be joint policies. Then, for any state $s$, we obtain the following expression:
\begin{align}
\mathrm{KL}(\boldsymbol{\pi}(\cdot \mid s), \overline{\boldsymbol{\pi}}(\cdot \mid s))&=\mathbb{E}_{\mathbf{a} \sim \pi}[\log \pi(\mathbf{a} \mid s)-\log \overline{\boldsymbol{\pi}}(\mathbf{a} \mid s)] \nonumber \\
& =\mathbb{E}_{\mathbf{a} \sim \pi}\left[\log \left(\prod_{i=1}^n \pi^i\left(\mathrm{a}^i \mid s\right)\right)-\log \left(\prod_{i=1}^n \bar{\pi}^i\left(\mathrm{a}^i \mid s\right)\right]\right] \nonumber \\
& =\mathbb{E}_{\mathbf{a} \sim \pi}\left[\sum_{i=1}^n \log \pi^i\left(\mathrm{a}^i \mid s\right)-\sum_{i=1}^n \log \bar{\pi}^i\left(\mathrm{a}^i \mid s\right)\right] \nonumber \\
& =\sum_{i=1}^n \mathbb{E}_{\mathbf{a}^i \sim \pi^i, \mathbf{a}^{-i} \sim \pi^{-i}}\left[\log \pi^i\left(\mathrm{a}^i \mid s\right)-\log \bar{\pi}^i\left(\mathrm{a}^i \mid s\right)\right] \nonumber \\
 &=\sum_{i=1}^n \mathrm{KL}\left(\pi^i(\cdot \mid s), \bar{\pi}^i(\cdot \mid s)\right)  = \sum_{i=1}^n \mathrm{KL}\left(\pi^i, \bar{\pi}^i\right).
\end{align}

Now, we have decomposed the terms of the MD update rule to derive the decomposition of the joint policy update rule for $\boldsymbol{\pi}$. The update rule is as follows:

\begin{align}
\boldsymbol{\pi}_{k+1} &\leftarrow\underset{\overline{\boldsymbol{\pi}} \in \boldsymbol{\Pi}}{\arg \max }  \;\mathbb{E}_{\mathrm{s} \sim \rho_{\boldsymbol{\pi}_{k}}, \mathbf{a} \sim \overline{\boldsymbol{\pi}}}\left[\sum_{m=1}^n A_{\pi_{k}}^{i_m}\left(\mathrm{~s}, \mathbf{a}^{i_{1: m-1}}, \mathrm{a}^{i_m}\right)\right]-\frac{1}{t_{k}} \sum_{i=1}^n \mathrm{KL}\left(\pi^i, \bar{\pi}^i\right) \nonumber \\
& =\underset{\overline{\boldsymbol{\pi}} \in \boldsymbol{\Pi}}{\arg \max }  \;\sum_{m=1}^n\mathbb{E}_{\mathrm{s} \sim \rho_{\boldsymbol{\pi}_{k}}, \mathbf{a} \sim \overline{\boldsymbol{\pi}}}\left[ A_{\pi_{k}}^{i_m}\left(\mathrm{~s}, \mathbf{a}^{i_{1: m-1}}, \mathrm{a}^{i_m}\right)-\frac{1}{t_{k}} \mathrm{KL}\left(\bar{\pi}^{i_m}, \pi_{k}^{i_m}\right)\right].
\end{align}

Based on the above summation, the agents' policies $\pi^{i_{1:n}}$ can be updated sequentially. We can formulate the update rule for each agent, e.g., $i_m$, as follows and employ multiple SGD steps at each iteration to approximately update the agent's policy:

\begin{align} 
\pi^{i_m}_{k+1} \leftarrow \underset{\pi^{i_m} \in \Pi}{\arg \max } & \; \mathbb{E}_{s\sim\rho_{\boldsymbol{\pi}_{k}}}\left[\mathbb{E}_{\mathrm{\boldsymbol{a}}^{i_{1:m-1}}\sim\boldsymbol{\pi}_{k+1}^{i_{1:m-1}},\mathrm{a}^{i_m}\sim\pi^{i_m}}  \left[A_{\boldsymbol{\pi}_{k}}^{i_m}\left(\mathrm{s},\mathrm{\boldsymbol{a}}^{i_{1:m-1}},\mathrm{a}^{i_m}\right)\right] \right] \nonumber \\
 &  - \mathbb{E}_{s\sim\rho_{\boldsymbol{\pi}_{k}}}\left[\frac{1}{t_{k}} \mathrm{KL}\left(\pi^{i_{m}}(\cdot \mid \mathrm{s}),\pi_{k}^{i_{m}}(\cdot \mid \mathrm{s}) \right)\right]. \label{eq:2}
\end{align}

In the above formula, when updating agent $i_m$, we consider the updated actions of the previously updated agents, denoted as  $\mathrm{\boldsymbol{a}}^{i_{1:m-1}}\sim\boldsymbol{\pi}_{k+1}^{i_{1:m-1}}$. This ensures that the joint policy reflects the latest actions of the other agents, enabling the agent $i_m$ to make informed updates based on the updated joint policy. However, recalculating the joint advantage function with previously updated agents can introduce computational overhead. To mitigate this, we can utilize the joint advantage estimator introduced in \citep{kuba2022trust} within our algorithm. This estimator allows us to estimate the joint advantage function for every state $s$ as follows:

\begin{align} 
\mathbb{E}_{\mathrm{\boldsymbol{a}}^{i_{1:m-1}}\sim\boldsymbol{\bar{\pi}}^{i_{1:m-1}},\mathrm{a}^{i_m}\sim\hat{\pi}^{i_m}}&\left[A_{\boldsymbol{\pi}}^{i_m}\left(\mathrm{s},\mathrm{\boldsymbol{a}}^{i_{1:m-1}},\mathrm{a}^{i_m}\right)\right] \nonumber \\ 
&=\mathbb{E}_{\mathbf{a} \sim \boldsymbol{\pi}}\left[\left(\frac{\hat{\pi}^{i_{m}}\left(\mathrm{a}^{i_{m}} \mid s\right)}{\pi^{i_{m}}\left(\mathrm{a}^{i_{m}} \mid s\right)}-1\right) \frac{\boldsymbol{\bar{\pi}}^{i_{1: m-1}}\left(\mathbf{a}^{i_{1: m-1}} \mid s\right)}{\boldsymbol{\pi}^{i_{1: m-1}}\left(\mathbf{a}^{i_{1: m-1}} \mid s\right)} A_{\boldsymbol{\pi}}(s, \mathbf{a})\right] .
\end{align}
where $\boldsymbol{\hat{\pi}}$ represent the policy of agent $i_m$, and $\boldsymbol{\bar{\pi}}$ denote the joint policy of agents $i_{1:m-1}$. We define $ M^{i_{1: m}}(s, \mathbf{a})=\frac{\boldsymbol{\pi}^{i_{1: m-1}}_{\theta_{k+1}}\left(\mathbf{a}^{i_{1: m-1}} \mid s\right)}{\boldsymbol{\pi}^{i_{1: m-1}}_{\theta_{k}}\left(\mathbf{a}^{i_{1: m-1}} \mid s\right)} A(s, \mathbf{a})$, where $A(s, \mathbf{a})$ represents the advantage function. By substituting this joint advantage estimator into Equation (\ref{eq:2}), we derive the HAMDPO loss function, which is as follows:
\begin{align} 
L^{i_m}(\theta,\theta_k) &= \mathbb{E}_{s\sim\rho_{\boldsymbol{\pi}_{\theta_{k}}},\mathbf{a}\sim\boldsymbol{\pi}_{\theta_{k}}}\left[\frac{\pi_{\theta}^{i_{m}}\left(\mathrm{a}^{i_{m}} \mid s\right)}{\pi_{\theta_{k}}^{i_{m}}\left(\mathrm{a}^{i_{m}} \mid s\right)} M^{i_{1: m}}(s, \mathbf{a}) \right] \nonumber \\
 & \, - \mathbb{E}_{s\sim\rho_{\boldsymbol{\pi}_{\theta_{k}}}}\left[\frac{1}{t_{k}} \mathrm{KL}\left(\pi_{\theta}^{i_{m}}(\cdot \mid \mathrm{s}),\pi_{\theta_{k}}^{i_{m}}(\cdot \mid \mathrm{s}) \right)\right], \label{eq:4}
\end{align}

At each iteration, we can apply the MD-style update rule for an arbitrary permutation of agents, such as drawing a shuffled permutation. In summary, the HAMDPO process we have introduced can be outlined using Algorithm \ref{alg:HAMDPO}. This algorithm provides a step-by-step description of the HAMDPO method.
\begin{algorithm}[H]
		\caption{Multi-Agent Mirror Descent Policy Optimization}
		\label{alg:HAMDPO}
		\begin{algorithmic}[1]
			\State Initialize the joint policy $\boldsymbol{\pi}_0=\left(\pi_0^1, \ldots, \pi_0^n\right)$.
			\For{$k = 0,1,2,...$ }
			\State Compute the advantage function $A_{\pi_k}(s, \boldsymbol{a})$ for all state-(joint)action pairs $(s, \boldsymbol{a})$. 
			\State Set $M^{i_{1}}(\mathrm{~s}, \mathbf{a})=A(\mathrm{~s}, \mathbf{a})$.
			\State Draw a permutation $i_{1: n}$ of agents at random. 
			\For{$m = 1:n$ }
			\State  Update actor $i^{m}$ with $g$ steps of SGD on the objective function given by Equation (\ref{eq:4}):
$$ 
  \begin{array}{l}
  	\theta_{k}^{(0)}=\theta_{k}, \quad \text { for } \quad i=0, \ldots, g-1 \\
  	\theta_{k}^{(i+1)} \leftarrow \theta_{k}^{(i)}+\left.\eta \nabla_{\theta} L \left(\theta, \theta_{k}\right)\right|_{\theta=\theta_{k}^{(i)}}, \\ \theta_{k+1}=\theta_{k}^{(g)}.
  \end{array}
  $$
			\State Compute $M^{i_{1: m+1}}(\mathrm{~s}, \mathbf{a})=\frac{\pi_{k+1}^{i_{m}}\left(\mathrm{a}^{i_{m}} \mid \mathrm{a}^{i_{m}}\right)}{\pi_{k}^{i_{m}}\left(\mathrm{a}^{i m} \mid \mathrm{a}^{i_{m}}\right)} M^{i_{1: m}}(\mathrm{~s}, \mathbf{a}) .$
			\EndFor
			\EndFor
		\end{algorithmic}
	\end{algorithm}
Our HAMDPO algorithm is similar to CPI, but it utilizes MD theory to replace the computation of maximum KL with mean KL, making it a more practical approach. Additionally, the step-size of HAMDPO is based on MD theory, leading to faster learning. HAMDPO also has connections to HATRPO and HAPPO algorithms.

Compared to HATRPO, HAMDPO does not explicitly enforce the trust-region constraint. Instead, it approximately satisfies it by performing multiple steps of SGD on the objective function of the optimization problem in the MD-style update rule. HAMDPO uses simple SGD instead of natural gradient, which eliminates the need to deal with computational overhead. In addition, the direction of KL in HAMDPO is consistent with that in the MD update rule in convex optimization, while it differs from that in HATRPO. Finally, HAMDPO employs a simple schedule motivated by the theory of MD, whereas HATRPO uses multiple heuristics to define the step-size and reduce it in case the trust-region constraint is violated.

HAMDPO and HAPPO both take multiple steps of stochastic gradient descent on unconstrained optimization problems, but they differ in how they handle the trust-region constraint. Recent studies, including \citealp{Engstrom2020}, suggest that most of the performance improvements observed in PPO are due to code-level optimization techniques. These studies also indicate that PPO's clipping technique does not prevent policy ratios from going out of bounds; it only reduces their probability. Consequently, despite using clipping, PPO does not always ensure that the trust-region constraint is satisfied.

\section{Experiments} \label{sec_Experiments}
\begin{wrapfigure}{r}{0.4\textwidth}
    \includegraphics[width=0.4\textwidth,trim= .2cm .2cm 1cm 1.45cm,clip]{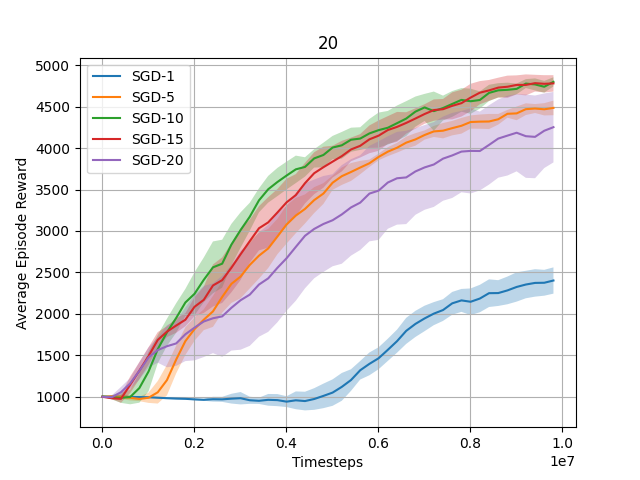}
  \caption{Effect of different values of SGD steps on the performance of HAMDPO algorithm for the Ant-v2 2x4 task.}
  \label{fig:sgd}
\end{wrapfigure}

This section presents an evaluation of our proposed algorithm, HAMDPO, on several tasks from the Multi-Agent Mujoco continuous benchmark and a map from StarCraftII. We compare the results obtained from HAMDPO with those obtained from HATRPO and HAPPO. The Multi-Agent Mujoco benchmark consists of a variety of robotic control tasks with continuous state-action space, where multiple agents collaborate to solve a task as distinct parts of a single robot. In contrast, StarCraftII has a discrete action space and is a game where each learning agent controls a specific army unit while the built-in AI controls the enemy units.

\begin{figure}
  \centering
  \includegraphics[width=1\textwidth]{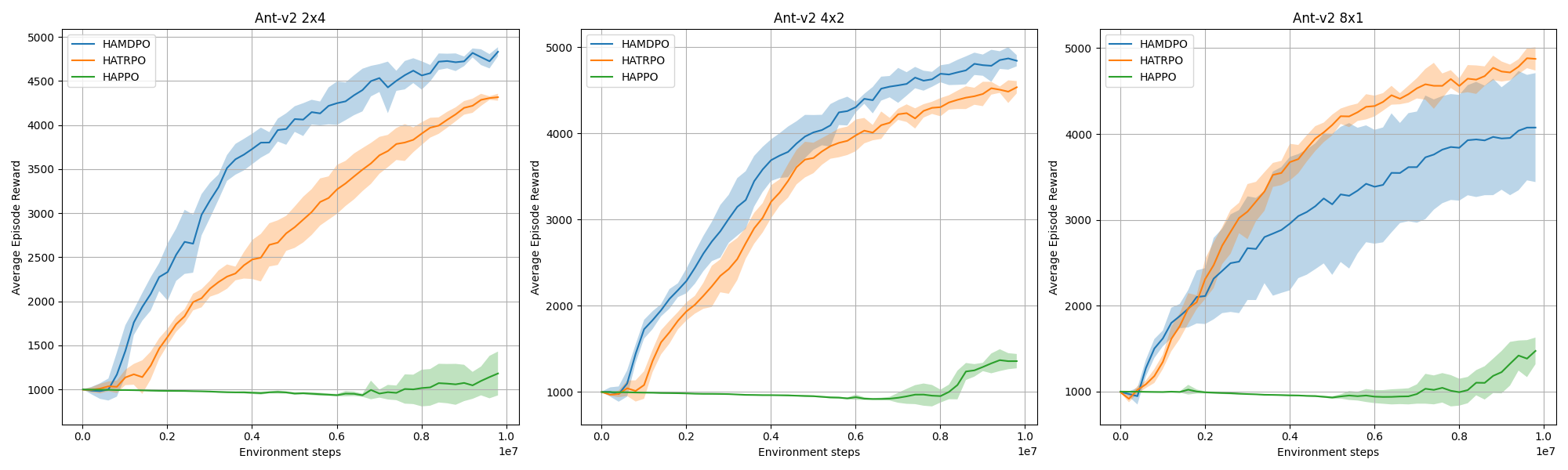}
  \caption{Comparison of average episode reward for HAMDPO, HATRPO, and HAPPO on three tasks of Ant-v2.}
  \label{fig:mujoco1}
\end{figure}

\begin{figure}
  \centering
  \includegraphics[width=1\textwidth]{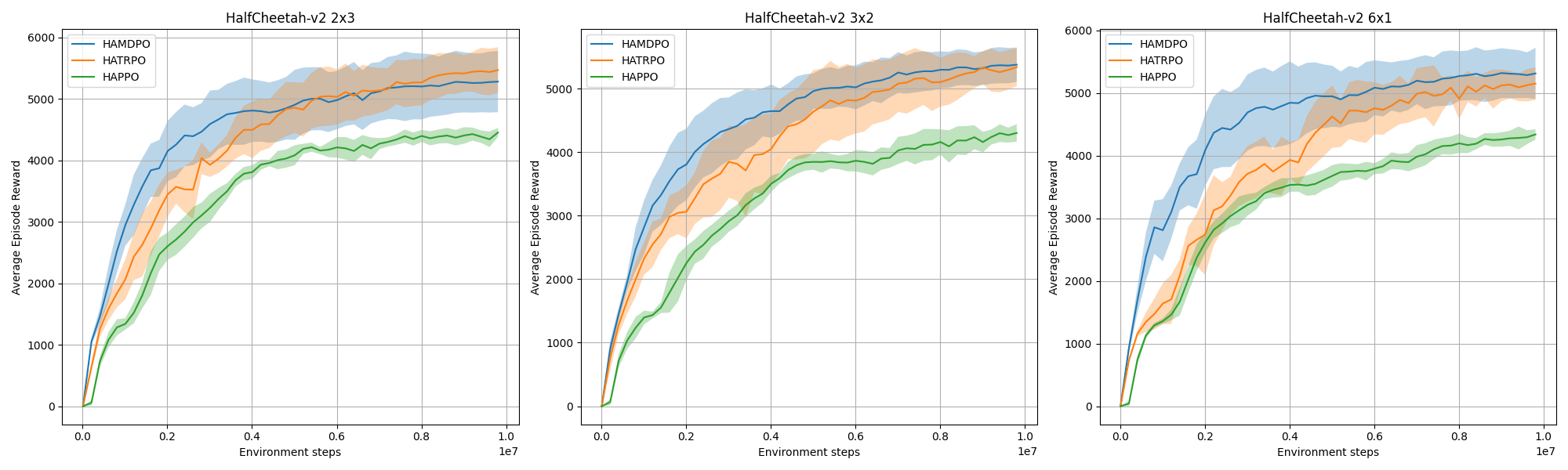}
  \caption{Comparison of average episode reward for HAMDPO, HATRPO, and HAPPO on three tasks of HalfCheetah-v2.}
  \label{fig:mujoco2}
\end{figure}

\begin{figure}
  \centering
  \includegraphics[width=1\textwidth]{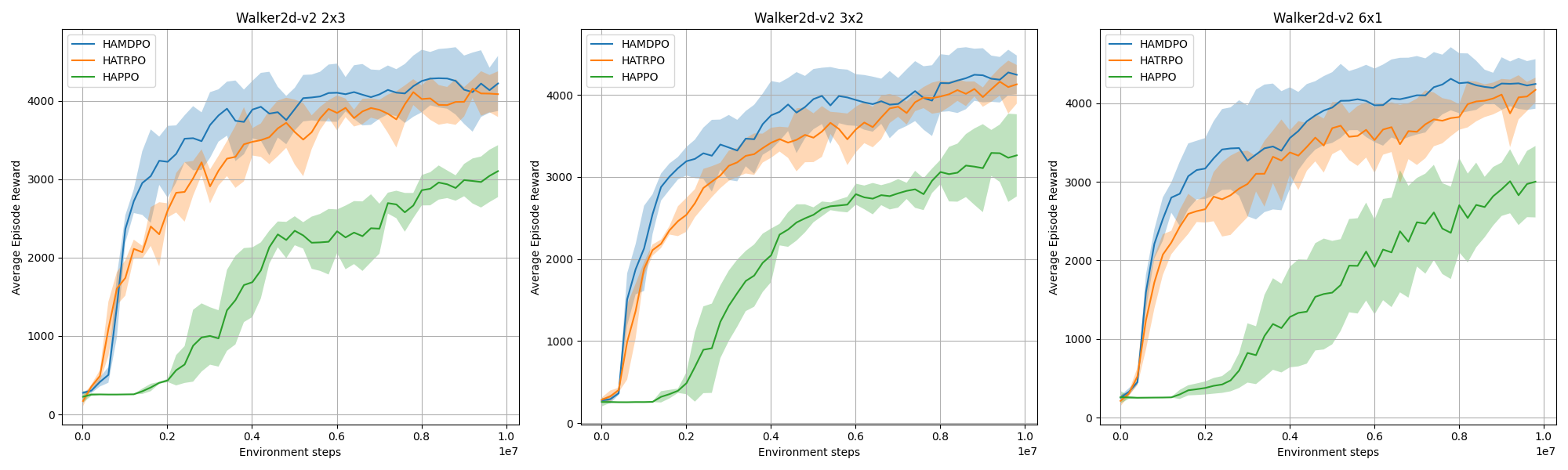}
  \caption{Comparison of average episode reward for HAMDPO, HATRPO, and HAPPO on three tasks of Walker2d-v2.}
  \label{fig:mujoco3}
\end{figure}

\begin{figure}
  \centering
  \includegraphics[width=0.6\textwidth]{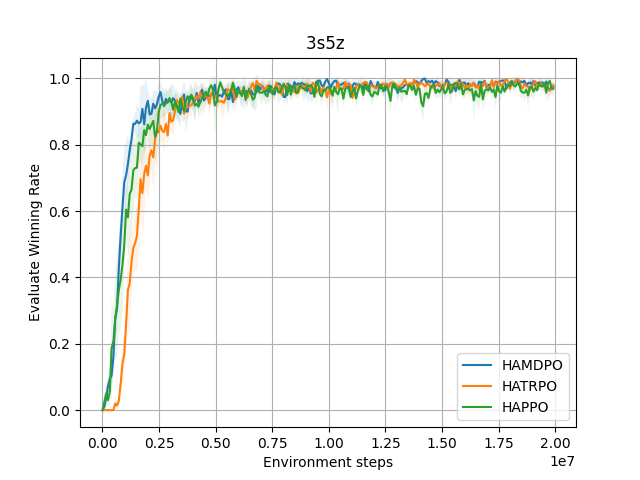}
  \caption{Comparison of evaluation win rates for HAMDPO, HATRPO and HAPPO on 3s5z map of StarCraft II. The mean win rates across five different runs with $95\%$ confidence intervals are shown.}
     \label{fig:smac}
\end{figure}

To measure the performance of our algorithm, we plot the average episode reward for multiple tasks of Multi-Agent Mujoco in Figures~\ref{fig:mujoco1}, \ref{fig:mujoco2}, and \ref{fig:mujoco3}
, and the mean evaluation winning rate of StarCraftII with a $95\%$ confidence interval across five different runs in Figure \ref{fig:smac} over $10^7$ time-steps\footnote{The code is available at \url{https://github.com/mehdinasiri/Mirror-Descent-in-MARL}}. The results demonstrate that HAMDPO outperforms both HATRPO and HAPPO in both continuous and discrete benchmarks.

Additionally, we examined the impact of the number of SGD steps per iteration on the Ant-v2 2x4 task, as presented in Figure \ref{fig:sgd}. Our findings reveal that vanilla policy gradient is obtained with one SGD step, whereas the regularization term becomes more relevant with more steps. We conducted HAMDPO with ten SGD steps for all the experiments. However, the outcomes depicted in Figure \ref{fig:sgd} suggest that significant performance can still be achieved with fewer steps. Nonetheless, it is crucial to acknowledge that using more steps to optimize will increase the duration of the optimization process.

\section{Conclusion}
In conclusion, this study introduced the HAMDPO algorithm, which is an innovative on-policy trust-region algorithm for cooperative MARL based on the theory of Mirror Descent and a theoretically-justified trust region learning framework in MARL. HAMDPO addresses challenges that arise in cooperative MARL settings to find optimal policies where agents have varying abilities and individual policies.

The HAMDPO algorithm updates agent policies iteratively by solving trust-region optimization problems that ensure stability and improve convergence rates. However, instead of solving these problems directly, the policies are updated by taking multiple gradient steps on the objective function of these problems. The experimental results demonstrate that HAMDPO outperforms state-of-the-art algorithms such as HATRPO and HAPPO.

Looking ahead, there are several potential future directions for this research. One direction is to develop an off-policy version of the HAMDPO algorithm in MARL. Additionally, further research could explore the potential of HAMDPO in large-scale scenarios with a large number of agents. Furthermore, it would be interesting to investigate the potential of Mirror Descent in other areas of MARL, such as in the context of competitive multi-agent settings.

\bibliography{sn-bibliography}

\end{document}